\def\year{2020}\relax
\documentclass[letterpaper]{article} 
\usepackage{aaai20_alter}  
\usepackage{times}  
\usepackage{helvet} 
\usepackage{courier}  
\usepackage[hyphens]{url}  
\usepackage{graphicx} 

\usepackage{dsfont}
\usepackage{amsmath}
\usepackage{algorithm}
\usepackage{algpseudocode}
\usepackage{amssymb}
\usepackage{mathtools}
\usepackage{multirow}
\usepackage{booktabs}

\DeclarePairedDelimiter{\ceil}{\lceil}{\rceil}
\newcommand{\norm}[1]{\left\lVert#1\right\rVert}

\algnewcommand\algorithmicforeach{\textbf{for each}}
\algdef{S}[FOR]{ForEach}[1]{\algorithmicforeach\ #1\ \algorithmicdo}

\DeclareMathOperator*{\argmax}{argmax} 

\usepackage{graphicx}  
\title{Neighborhood Watch: Representation Learning with Local-Margin Triplet Loss and Sampling Strategy for K-Nearest-Neighbor Image Classification}
\author{Phawis Thammasorn\textsuperscript{\rm 1} Daniel Hippe \textsuperscript{\rm 3} Wanpracha Chaovalitwongse \textsuperscript{\rm 2} Matthew Spraker \textsuperscript{\rm 4} \\ \Large \textbf{ Landon Wootton \textsuperscript{\rm 5} Matthew Nyflot \textsuperscript{\rm 5} Stephanie Combs \textsuperscript{\rm 6} Jan Peeken \textsuperscript{\rm 6} Eric Ford \textsuperscript{\rm 5}} \\
\textsuperscript{\rm 1} Department of Industrial Engineering, University of Arkansas, AR \\
pthammas@uark.edu 
\textsuperscript{\rm 2} Department of Industrial Engineering, University of Arkansas, AR \\
\textsuperscript{\rm 3} Department of Radiology, University of Washington, WA \\
\textsuperscript{\rm 4} Department of Radiation Oncology, Washington University, MO \\
\textsuperscript{\rm 5} Department of Radiation Oncology, University of Washington \\
\textsuperscript{\rm 6} Department of Radiation Oncology, Klinikum rechts der Isar, Technical University of Munich (TUM) \\
}
 
\begin{document}

\maketitle

\begin{abstract}
Deep representation learning using triplet network for classification suffers from a lack of theoretical foundation and difficulty in tuning both the network and classifiers for performance. To address the problem, local-margin triplet loss along with local positive and negative mining strategy is proposed with theory on how the strategy integrate nearest-neighbor hyper-parameter with triplet learning to increase subsequent classification performance. Results in experiments with 2 public datasets, MNIST and Cifar-10, and 2 small medical image datasets demonstrate that proposed strategy outperforms end-to-end softmax and typical triplet loss in settings without data augmentation while maintaining utility of transferable feature for related tasks. The method serves as a good performance baseline where end-to-end methods encounter difficulties such as small sample data with limited allowable data augmentation.
\end{abstract}
\section{Introduction}
Deep representation learning using Triplet Network has recently been successful in various computer vision tasks \cite{gordo2016deep,Yan2018CVPR}. Originally proposed by \citeauthor{hoffer2015deep}, the architecture consist of three identical networks dedicated for extracting embedded feature of different types relative to anchor sample for training with some metric criteria and transferring to subsequent tasks. 

Other than prominent successes of the network in image retrieval and matching, there are multiple works that sought to adapt the network for image classification tasks \cite{simo2016fashion,liu2017scene}. Commonly used loss criteria, such as \cite{balntas2016learning}, train the network using a distance metric between intra-class and inter-class samples such that same-class features are clustered together away from that of the different-class. General intuition is that the separation in feature space improves subsequent classifier for good performance. Outside of the general image domain, uses of the network are also explored in medical image classification and analysis \cite{codella2018collaborative,nyflot2019deep}. Some successes in the domain suggest potential of the network in coping with small and imbalanced data due to the fact that triplet sampling creates larger training instances consisting of positive and negative samples while simultaneously training with both classes in every iteration \cite{zhang2018similarity,thammasorn2018deep}. 

Despite multiple successes, the strategy introduces difficulty regarding performance validation. Classification performance depends on effective training and validation of both the embedding network and the classifier, which are typically done independently such that loss calculation of the extractor has no relation to tuning of the classifier. Although class-separation criteria is appealing in general, it doesn't theoretically guarantee better performance for a given classifier. It is also hard to determine whether inferior performance whether is caused by poor embedding or the classifier. Some works in literature suggest combining triplet loss with classification loss, such as softmax, and training the network in a unified end-to-end manner,
it is not a universal approach as some applications, especially in the medical image domain, may require transparent and interpretable models \cite{peng2019multi} and prefer transfer learning approaches. 

To address the problem, we propose a modification to the triplet network training strategy such that training the network would theoretically result in better performance when feature is transferred for pre-set nearest-neighbor classification. Our contributions proposed in this paper are three-fold:
\begin{itemize}
\item A novel loss function and training strategy that integrates hyper-parameter information of a nearest neighbor model such that minimizing the loss results in suitable embedding for subsequent classification.
\item New insights into typical triplet loss gain through the perspective of a nearest-neighbor model and demonstration of better performance through our appraoch.
\item Demonstration through experiments on the potential of the proposed strategy in transfer learning to other classification models and related tasks.\\
\end{itemize}
\section{Related works}
\subsection{Representation Learning using Triplet Network}
Given raw input data $\mathnormal{I}$, goal of representation learning is to learn suitable feature $\mathnormal{X=f(I)}$ for later operations. For Computer Vision tasks, a popular strategy is to use deep learning model pre-trained for large-scale classification such as ImageNet \cite{szegedy2017inception} as $\mathnormal{f(I))}$. Application along this strategy have been implemented in many domains and problems \cite{han2018new,baltruschat2018orientation}. even though the approach is applicable to general image, it requires the input data to be resized and fit input setting of the pretrained architecture which usually corrupt input images. 

Metric learning using is an alternative deep learning strategy for representation learning. Unlike the popular approach, it was proposed to learn feature embedding $\mathnormal{X=f(I; \theta)}$ where $\theta$ is trainable parameter tuned with some metric such that the resulting feature is suitable for later tasks after optimizing. Triplet Network is an architecture designed for the purpose. It can be conceptualized as organization of extractor and comparator networks \cite{nyflot2019deep}. Extractor network or embedding network digest input data sample into a feature vector. Choice of architecture for the extractor depends on the nature of the input data, such as a feed-forward neural network for vector data and a Convolutional Neural Network (CNN) for image data. Typically, three identical extractors with shared parameter $\theta$ simultaneously create feature vectors of anchor point input $I_a$, same-class (positive) input $I_p$, and different-class (negative) input $I_n$. The result features are $X_a, X_p,$ and $X_n$ correspond to respective input types. Then, the comparator network evaluates feature vectors and backpropagate error signal to adjust network parameters. \citeauthor{hoffer2015deep} proposed general idea for metric loss criteria such that features of same-class pairs are clustered together and feature of different-class pairs are well separated in multidimensional feature space. The idea was later simplified with fixed margin triplet loss
\begin{equation}
\mathcal{L}_{tri}=max\Big(0, D_{a,p}-D_{a,n}+m \Big)
\end{equation}
Where $\mathnormal{D_{a,p}}=\norm{X_a-X_p}^2$ is distance metric between an anchor feature and positive feature, $\mathnormal{D_{a,n}}=\norm{X_a-X_n}^2$ is distance between the anchor feature with a negative feature, and $\mathnormal{m}$ is fixed margin constant. Verbally, minimizing the loss mean the feature of different class must be separated at least $m$ distance which result in class-separated feature space. 

Training of the network is done from expanded dataset of triplet sample. Given $\mathnormal{I=}\{\mathnormal{I_1,I_2,...,I_n}\}$ and $\mathnormal{C=}\{\mathnormal{C_1,C_2,...,C_n}\}$ as input data and corresponding class label, the triplet training dataset is constructed as set of triplet $\mathnormal{T=(I_a,I_p,I_n)}$ where $\mathnormal{I_a}$ is a sample from the initial dataset of any class. \citeauthor{hermans2017defense} improves the loss with hard-positive and hard-negative mining strategy which is a triplet sampling strategy that choose $I_p$ corresponds to largest $D_{a,p}$ and $I_n$ corresponds to smallest $D_{a,p}$ as representation of a batch for loss calculation. Although proposed for image retrieval, the sampling is also used for classification task \cite{peng2019multi}.

After training, the comparator network is discarded. A single extractor is then designated for the embedding function $\mathnormal {f(I; \theta)}$. As an alternative to pre-trained approach, training with the expanded dataset potentially helps cope with the small size of the original dataset while reducing reliance on resizing data and data augmentation. Despite lack of theoretical foundation for good performance under small and imbalance data, progress along this line are explored in multiple works \cite{zhang2018similarity,thammasorn2018deep,nyflot2019deep}.

\subsection{K-Nearest-Neighbor classification}
K-Nearest Neighbor (KNN) is a classical machine learning approach to estimate local distribution of interested value $\mathnormal{p(C|X)}$. Specifically, given $\mathnormal{n}$ known pairs of feature sample, associate class label $\mathnormal{(X_1,C_1),(X_2,C_2),...(X_1,C_1),(X_n,C_n)}$, and query point with unknown value $\mathnormal{(X_q,C_q)}$, $\mathnormal{p(C_q|X_Q)}$ and $\mathnormal{C_q}$ can be estimated as 
\begin{equation}
p(C_q=c|X_Q)=\sum_{i\in\mathcal{N}_{q,d_{q,k}}}\frac{\mathds{1}(C_i=c)}{k}
\end{equation}
\begin{equation}
C_q=\argmax_c  p(C_q=c|X_Q)
\end{equation}
where $k$ is pre-defined hyper-parameter of how many nearest neighbor to be included in the estimation, set $\mathcal{N}_{q,k}$ contain index $\mathnormal{i}$ of which $\mathnormal{X}_{i}$ belongs to neighbors of $\mathnormal{X}_{q}$. Let $\mathnormal{d}_{q,k}$ be distance from $\mathnormal{X}_{q}$ to its $k$th nearest neighbor point. The classification considers that all samples covered in radius $\mathnormal{d}_{q,k}$ as neighbors of $\mathnormal{X}_{q}$, and belong to common local probability distribution $p(C_q|X_Q)$ which can be estimated with fraction of total occurrances. Notably, Larger $k$ value increase confidence in the estimation but also increase possibility that some samples in the neighborhood doesn't belong to the same distribution. On the other hand, smaller $k$ ensure that samples are less likely from different distribution but decrease confidence in the estimation. While any $k\leq n$ is possible, \citeauthor{duda2012pattern} posits that the estimation of $\mathnormal{(C_q|X_q)}$ converge to actual probability when neighborhood space grow sufficiently small as $\mathnormal{n}\rightarrow\infty$ (e.g $\mathnormal{k}_{q}=\sqrt{n}$).

According to \cite{duda2012pattern}, nearest neighbor (NN) classification is a sub-optimal method such that its highest error-rate is bounded at double of Bayes error rate, its non-parametric nature make it a classifier of choice especially in scenario where training set is inadequate for many parametric methods. Furthermore, estimation from the approach can be used to further analyze distribution of feature space and for probabilistic interpretation, especially with feature from deep learning models \cite{papernot2018deep,peng2019multi}.

Simplest implementation of nearest neighbor approach requires full-pass search on entire training dataset for each query point. However, there are many techniques such as kd-tree and ball-tree \cite{finley2008efficient,liaw2010fast} that greatly reduce accessing time in search for nearest neighbor samples. Its simplicity and solid theoretical foundation make NN approach a good baseline for machine learning problems.

We use both concepts of triplet metric learning and KNN classification in formulating propose method to create feature that can be used for classification or further analysis based on NN approach.

\section{Propose method}
We propose modification to triplet loss to ensure feature with better performance for subsequent KNN baseline with predefined value of $K$. To overcome difficulty from the independent tuning of the network and classifier, fixed margin triplet loss is reformulated using neighborhood concept with purpose to train Triplet network for feature with more homogenous local distribution such that error in KNN classification is less like if query point belong to same distribution within the feature space.  

There are two assumptions in formulating our method. 
\begin{itemize}
\item First, distribution in feature space change only slightly in every training iteration of triplet training such that negative and positive samples are incrementally pushed away and pulled in respect to an anchor point. 
\item Second, corresponding feature of query points belong to same data distribution in training feature space such that occurrence of outlier query, the samples which belong to subspace where there is less or no samples from the dataset, is not frequent and can be ignored. The assumption also implies that query feature is more likely to be correctly classified if probability of having same-class sample is larger than that of having different-class sample in common local subspace as query and known sample belong to same distribution. 
\end{itemize}
\subsection{Local margin and relaxed-boundary search}
We propose using distance to $\mathnormal{k}$th nearest positive neighbor to anchor point in triplet sample as margin for the triplet loss. The modification results in following triplet local-margin loss
\begin{equation}
\mathcal{L}_{lm}=max\Big(0, D_{a,p}-D_{a,n}+c_b\cdot d_{a,k+} + \epsilon \Big)
\end{equation}
where $\epsilon$ is small positive constant, $\mathnormal{d_{a,k+}}$ is distance to $\mathnormal{k}$th nearest positive neighbor from $\mathnormal{X_{a}}$, and $c_b$ is a constant of which $c_b\geq 3$. Verbally, minimizing the loss would push negative samples away neighborhood boundary of $\mathnormal{X_{a}}$ while pulling positive sample further inside. The value of $\mathnormal{k}$ is fixed throughout training.

Practically, it is inefficient to access for $\mathnormal{d_{a,k+}}$ in every update iteration as mentioned in \cite{hermans2017defense}. Thus, we relax the search problem by calculating and fixing value of $\mathnormal{d_{a,k+}}$ for each possible $\mathnormal{X_{a}}$ at the beginning of each training epoch. As stated in first assumption above, the fixed $\mathnormal{d_{a,k+}}$ serves as snapshot into of distance to neighborhood boundary of each $\mathnormal{X_{a}}$ which may be slightly different than actual boundary distance. Nevertheless, it is updated regularly as epoch progress and used improve distribution among its neighborhood. With the approach and existing data structure for KNN search, accessing dataset can be done efficiently and avoid adding too much time the training update.

The following propositions and theorem posit that that minimizing the local-margin loss improves performance of knn classifier.\\
\textbf{Proposition 1.1:} Given neighborhood defined by $\mathcal{N}_{a,d_{a,k}}$, statement $D_{a,i}\leq d_{a,k}$ is true for any known sample $\mathnormal{X_{i}}$ where $\mathnormal{i}\in\mathcal{N}_{a,d_{a,k}}$. Otherwise, $D_{a,i}> d_{a,k}$ and $\mathnormal{i}\notin\mathcal{N}_{a,d_{a,k}}$.\\
\textbf{Proposition 1.2:} Given a query sample with unknown class $\mathnormal{X_{q}}$ which has closest distance to known sample $\mathnormal{X_{a}}$, $\mathnormal{X_{q}}$ is either fall within neighborhood area cover by $d_{a,k}$ such that  $D_{a,q}<= d_{a,k}$, or an outlier such that $D_{a,q}>d_{a,k}$\\
In words, the statements define conditions of sample $\mathnormal{X_{i}}$ which fall within neighborhood of $\mathnormal{X_{a}}$ such that its coordinates is covered by radius around $\mathnormal{X_{a}}$ to be consider one of its neighbor. The same conditions also applied to query sample of unknown class which is subjected to testing in K-NN classification that is closest to $\mathnormal{X_{a}}$.\\
\textbf{Theorem 1:} If local-margin loss is minimized for all $\mathnormal{X_{a}}$ , then any unknown $\mathnormal{X_{q}}$ closest to $\mathnormal{X_{a}}$ and is not an outlier has a neighborhood $\mathcal{N}_{q,d_{q,k}}$ of which all sample $\mathnormal{X_{j}}\in\mathcal{N}_{q,d_{q,k}}$ are of class label as $\mathnormal{X_{a}}$. \\
Proof: Consider minimized $\mathcal{L}_{tri-lm}=0$. Then, $d_{a,k}=d_{a,k+}$ and the following statement can be rearranged from $\mathcal{L}_{tri-lm}$ definition and true for any $\mathnormal{X_{a}}$ with its possible $\mathnormal{X_{n}},\mathnormal{X_{p}}$
\begin{equation}
D_{a,n} \geq D_{a,p}+c_b\cdot d_{a,k}+\epsilon > 3\cdot d_{a,k}
\end{equation}
Consider that there are $k$ sample cover in radius $d_{a,k}$ away from $\mathnormal{X_{a}}$ plus one center sample $\mathnormal{X_{a}}$.
If the inequality above hold, then there are at least $k+1$ known samples with common class label covered by radius $ c_b\cdot d_{a,k}$ from $\mathnormal{X_{a}}$ as all the negative samples are further away from the radius. Let $R_{q,a,k}$ be distance from $\mathnormal{X_{q}}$ that covers all the $k$ neighbors of $\mathnormal{X_{a}}$. Then, according to triangle inequality the following statement is true
\begin{equation}
R_{q,a,k}\leq D_{a,q}+d_{a,k}
\end{equation}
\begin{figure}[t]
\begin{center}
\centerline{\includegraphics[width=0.6\columnwidth,scale=0.25]{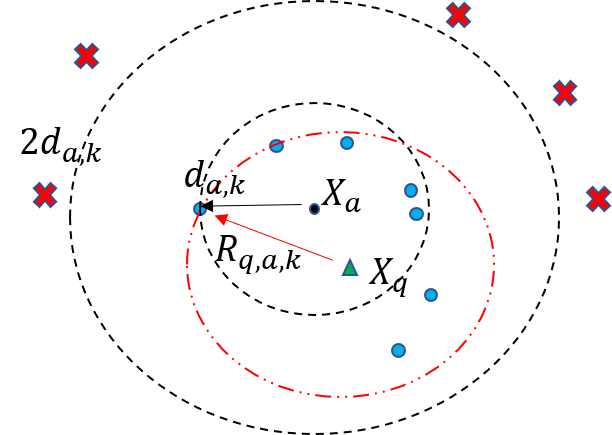}}
\caption{Example visualization of neighborhood surrounding $X_a$ and $X_q$.}
\label{FigRadius}
\end{center}
\end{figure}
Figure~\ref{FigRadius} illustrates the neighborhood concerning the inequality. Notice also that that area covered by $R_{q,a,k}$ from $\mathnormal{X_{q}}$ contain at least $k+1$ known samples. Thus, $d_{q,k}\leq R_{q,a,k}$ is true. Since proposition 1.2 states that $D_{a,q}\leq d_{a,k}$, $D_{a,q}$ in the inequality above can be substituted with $d_{a,k}$ and become 
\begin{equation}
d_{q,k}\leq R_{q,a,k}\leq 2\cdot d_{a,k}
\end{equation}
With above inequality, proving by contradiction can be done. Assuming that there are some $\mathnormal{X_{j}}\in\mathcal{N}_{q,d_{q,k}}$ belong to different class, then by proposition 1.1,1.2, and triangle inequality lead to the following
\begin{equation}
D_{a,n(j)}\leq D_{a,q}+D_{q,n(j)}\leq d_{a,k}+d_{q,k}+\leq 3\cdot d_{a,k}
\end{equation}
where n(j) denote the index that assume to be negative sample. Consider $D_{a,n(j)}$ as an instance of $D_{a,n}$, the statement become $D_{a,n}< c_b\cdot d_{a,k}+\epsilon$ which contradicts the stated optimal condition that $D_{a,n}\geq c_b\cdot d_{a,k}+\epsilon$. Thus, the contradiction proves the theorem.

The theorem implies that training to minimized $\mathcal{L}_{tri-lm}$ results in better performance of K-NN classification. Specifically, local distribution of class samples is becoming more homogenous as training create more neighborhood without different-class samples for non-outlier query points. Thus, K-NN can easier correctly classify according to the second assumption mention previously. The strategy also suggest setting fixed $k$ value according to desired sample support for the local estimations in both training and performance evaluation. Even if the loss is not fully minimized such that search for best $k$ need to be done, the value may be used as upper limit such that setting greater value would unnecessarily cover different-class samples in the neighborhood.\\
\textbf{Corollary 1:} Same optimal condition of neighborhood hold even if local-margin $c_b\cdot d_{a,k}$ in the loss function is replaced by any fixed constant $m>3\cdot d_{a,k}$\\
Prove by contradiction can be done by following steps of the proof in theorem 1 with $c_b\cdot d_{a,k}$ term replaced by $m$. Insight of the corollary is that it implies that feature trained using fixed margin triplet loss can still perform well with KNN classification model if it can ensure that the margin $m$ is greater than $3\cdot d_{a,k}$ for all possible $X_a$. However, typical triplet loss with fixed margin doesn't guarantee the case as $m$ is usually fixed arbitrarily. In contrary, the proposed local-margin loss make sure that the condition hold by keeping track of value of $d_{a,k}$ for all $X_a$. Nevertheless, the insight justifies general intuition of setting $m$ as large as possible although the margin doesn't help specifying number $k$ later.

\subsection{Local positive and local negative mining}
Training with triplet can be inefficient with normal triplet sampling as positive samples near anchor point and negative samples far away from it may be selected regularly. In constructing triplet sample, we address the problem by considering local positive and local negative sample inside neighborhood of $\mathnormal{X_{a}}$. Given $\mathnormal{X_{a}}$ and its neighborhood indices $\mathcal{N}_{a,d_{a,k}}$, we sample for corresponding $\mathnormal{X_{n},X_{p}}$ such that $n\in\mathcal{N}_{a,d_{a,k}}$ and $p\notin\mathcal{N}_{a,d_{a,k}}$. The sampling strategy means that the training push out known local negative sample away from the neighborhood while seeking known non-local positive to pull inside. The strategy is a relaxed version of hard-positive and hard-negative mining from \cite{hermans2017defense}. The proposed strategy doesn't require large batch size to ensure good quality of hard-positive and hard-negative samples. However, it suggests setting sizable $k$ such that more local negative can be sampled.

Nevertheless, the local mining strategy require keeping watch on available local positive and local negative for each $\mathnormal{X_{a}}$. We used similar relaxation to the proposed search strategy such that $\mathcal{N}_{a,d_{a,k}}$ is created for each $\mathnormal{X_{a}}$ at the beginning of each training epoch. With the first assumption, large portion of known local neighbors are still valid for the sampling strategy with a trade-off that some easy negative may be selected and some hard negative may be neglected.

\begin{algorithm}[t] \caption{Training with Local-margin Loss with Optional Local Positive/Negative Mining}
\textbf{Input:} $I\in\mathcal{I}$ raw training input, $e_{max}$ max \# of epoches, $k$ neighbors
\begin{algorithmic}[1]
\State Initilize $\theta,e=0$
\Repeat
\State $\mathcal{X} \gets f(\mathcal{I};\theta)$
\State $\mathcal{M} \gets$ NN model with $k$ and $\mathcal{X}$ 
\ForEach {$X_a \in \mathcal{X} $}
\If{ Local Mining }
\State $\mathcal{N}_{a,d_{a,k}} \gets \mathcal{M}.neighbors(X_a)$
\State $T \gets$ Triplet sampling $(X_a,X_n,X_p)$\\ where   $n\in\mathcal{N}_{a,d_{a,k}}, p\notin\mathcal{N}_{a,d_{a,k}}$
\Else
\State $T \gets$ Triplet sampling $(X_a,X_n,X_p)$\\ where $X_n\in\mathcal{X}, X_p\in\mathcal{X}$
\EndIf
\State $\Delta\mathcal{L}\gets minimizeLoss(T,\theta)$
\State $e \gets e+1$
\EndFor
\Until {$\Delta\mathcal{L} < \epsilon$ or $e> e_{max}$}
\end{algorithmic}
\end{algorithm}
Training with proposed triplet loss is summarized in algorithm 1. In each epoch, the algorithm start by extracting feature from set of training input and creating NN data structure model for searching of $k$th NN and $d_{a,k}$. For each sample in the dataset, corresponding positive and negative sample are collected to create Triplet samples which are used for loss calculation. The calculated loss is the used for training $\theta$.

According to proposed theorem 1 and corollary 1, it is beneficial to keep value of $d_{a,k}$ small such that smaller loss has to be reduced. A simple way in doing so is to apply global loss term as proposed in \cite{kumar2016learning} for regularize value of distances between feature samples. Thus, we formulate our loss function as
\begin{flalign*}
\mathcal{L}= w_{lm}\sum_{a}\sum_{p}\sum_{n} max(0, D_{a,p}-D_{a,n}+c_b\cdot d_{a,k}+\epsilon)&\\\ +\ w_{ms}\mu_s-\ w_{md}\mu_d\ +\ w_{ss}{\sigma^2}_s\ + \ w_{sd}{\sigma^2}_s
\end{flalign*}
where $\mu_s=\mathrm{E}[D_{a,p}], \mu_d=\mathrm{E}[D_{a,n}], {\sigma^2}_s$ is variance of $D_{a,p}, {\sigma^2}_d$ is variance of $D_{a,n}, \{w_{lm},w_{ms},w_{md},w_{ss},w_{sd}\}$ is set of weight constants. $p\notin\mathcal{N}_{a,d_{a,k}}$ and $n\in\mathcal{N}_{a,d_k}$ if local positive and negative mining is performed. The propose loss reduce local margin loss, reduce average same-class distance, and reduce variance terms while increasing average different-class distance. Small $\mu_s$ and local-margin term imply that value of $d_{a,k}$ is smaller as the loss reduced. Also regularizing variance terms potentially reduce number of outlier as it keeps feature not to scatter too far apart. $mu_d$ term prevent feature value from collapsing to zero.

\section{Experiments}
The proposed method is evaluated in 4 classifications tasks, namely digits recognition, object recognition, Epid-radiomics error classification, and Sarcoma survival prediction. The following are detail of the experiments.
\subsection{Dataset}
\begin{figure}[t]
\begin{center}
\centerline{\includegraphics[width=1.0\columnwidth,scale=0.5]{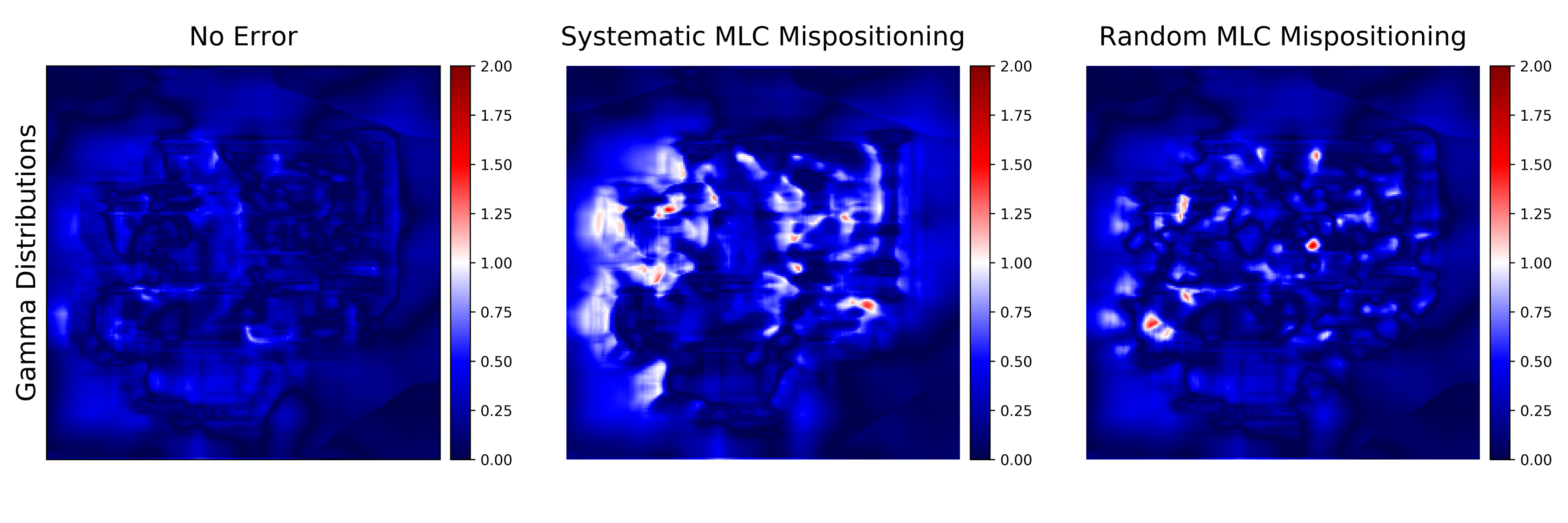}}
\caption{Example of Gamma images generated from EPID images. From left to right, image of no-error class, systematic-error class, and random-error class}
\label{NewFigDPI}
\end{center}
\end{figure}

\begin{figure}[t]
\begin{center}
\centerline{\includegraphics[width=\columnwidth]{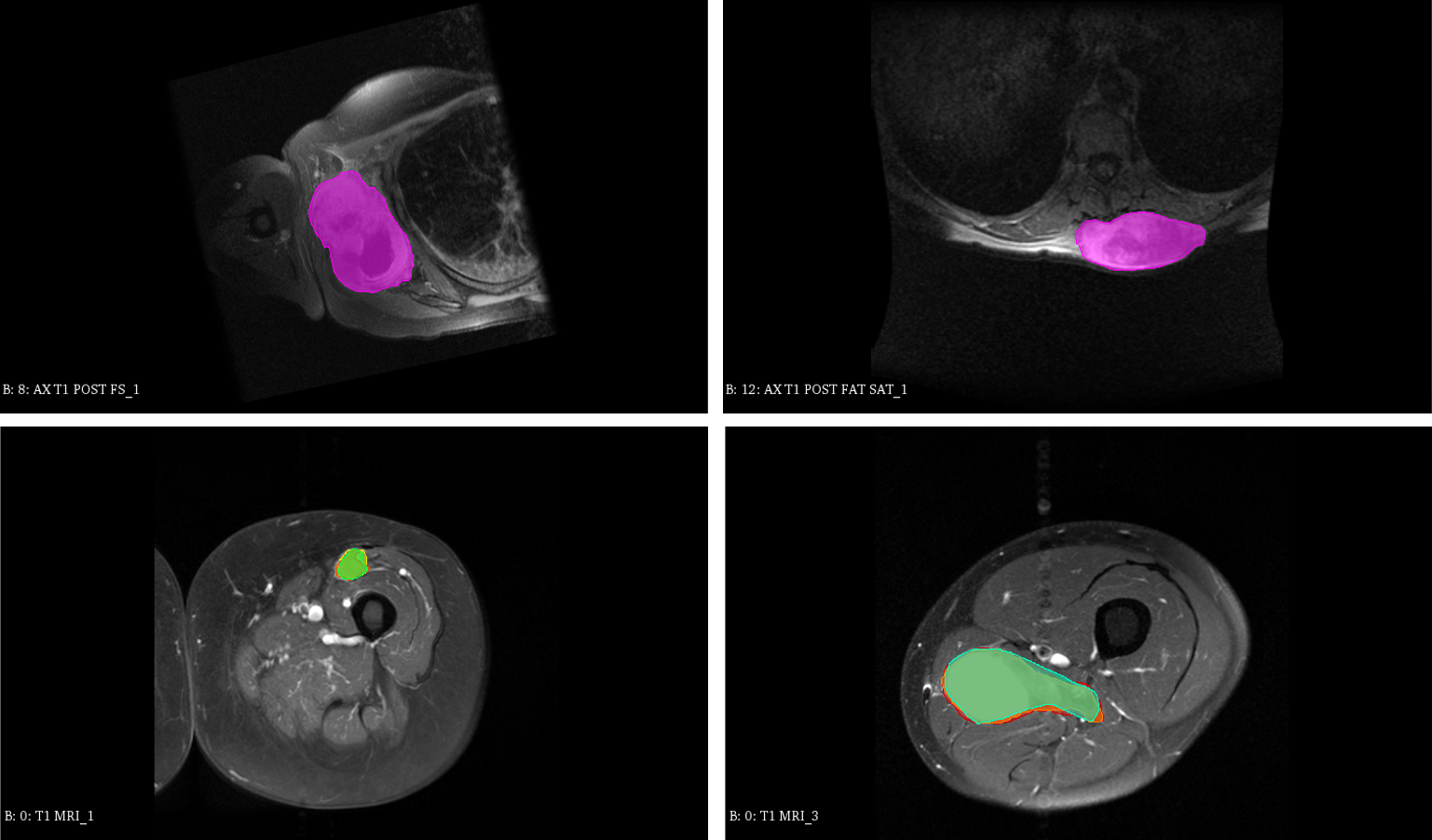}}
\caption{Example visualization of sample in MRI dataset with segmented ROI defined by expert. Top: samples from cohort1 patients. Bottom: samples from cohort2 patients.}
\label{Visualization}
\end{center}
\end{figure}

The following are 4 datasets used in evaluation of propose method for classification task. Two of the datasets are sizable public benchmark for images classification while the other are small real-world datasets developed in-house for medical imaging research.\\
\textbf{MNIST} dataset \cite{lecun-mnisthandwrittendigit-2010} consists of 28x28 gray-scale images of hand-written digits of which corresponding class labels are number from 0 to 9. The dataset is separated into 10,000 images for testing, 54,000 images for training, and 6,000 images for validation randomly sampled from original training set.\\
\textbf{Cifar-10} dataset \cite{krizhevsky2009learning} consists of 32x32 color images 10-classes real-world objects. The dataset is separated into 10,000 images for testing, 45,000 images for training, and 5,000 images for validation randomly sampled from original training set.\\
\textbf{EPID simulated Gamma images} \cite{nyflot2019deep} is set of 2-d images collected for dose measurement for radiation therapy. In radiation therapy delivery, the electronic portal imaging device (EPID) is used to capture the radiation beam to form 2-d images. The images then undergo gamma analysis to determine the safety and quality of the radiation treatment delivery by trained personnel. The goal of the classification task is to classify the images whether there is error in the delivery when known error is introduced. The dataset contains 558 Gamma (256x256) images from simulation with 186 intensity-modulated radiation therapy (IMRT) beams from 23 IMRT plans. Each beam has 3 classes of image namely, no-error, random multileaf collimator (MLC) mispositioning error (random-error), and systematic MLC mispositioning error (systematic-error). Figure~\ref{NewFigDPI} shows examples from each class. The images are separated into 255 images for testing, 273 images for training, and 30 images for validation randomly sampled from original 303 training images.  \\
\textbf{3D MRI Scan of Sarcoma Tumors} \cite{spraker2019mri} are sets of magnetic resonance imaging (MRI) scans of patients with sarcoma soft-tissue cancer. Clinical outcome were recorded including time-to-death censored from the day of scan acquisition and diagnosis. The goal is to classify whether the patients would survive longer than 1096 days (~3 years) after diagnosis. In this study, censored patients due to loss of followup are considered that they survived longer than 1096 days. Pre-treatment contrast-enhanced T1-weighted 3D MRI scans are acquired from two independent cohorts of patients diagnosed with biopsy-proven soft tissue Sarcoma (STS) from two different institutes of censored institute1 (cohort1) and censored institute2 (cohort2).  The acquisition is done with institutional picture archiving and communication system (PACS) standard with similar image matrix and resolutions. All patients that were less than 18 years old or were diagnosed with Kaposi or primary bone sarcomas were excluded. Included patients had sarcomas of various histologies of the extremity, trunk, or retroperitoneum. This study focused on American Joint Committee on Cancer (AJCC) version 7 stage II-III patients only, which encompasses non-metastatic patients with large (i.e. $>5$ cm) and/or higher grade (i.e. $>1$) tumors. Patients with image artifacts due to multiple MRI acquisitions within the tumor area were also excluded. The total number of patients in two cohorts were 200 and 72 patients, respectively. First 200 patients from cohort 1 were randomly separated into a training set of 180 images validation set of 20 images. Patients in cohort 2 were used for testing set. In all datasets, expert radiation oncologista evaluated each image for quality and manually segmented the gross tumor as ROI. Figure~\ref{Visualization} visualize some sample in our datasets. Each scan's ROI is resampled to have fixed resolution of 1 $mm^3$. All image ROIs are then resampled again into $64\times64\times64$ voxels.

\subsection{Setting}
We evaluate our method with 1 classification and 2 transfer tasks. The first experiment is comparison of classification accuracy testing all datasets which evaluate capability of features after training with our proposed method. Different extractor networks are designated for each dataset. In all architectures, leaky rectified linear unit (l-ReLu) activation defined as $max(0.01x,x)$ is used in all convolutions and feed-forward layers. For MNIST dataset, simple CNN architecture with 2 convolution layers each followed by max-pooling. The filter size is 3x3x32 and 3x3x64 respectively. The data after convolution and max-pooling is flatten and sent to feed-forward layer with output dimension of 128. For Cifar-10, we follow transfer learning approach in \cite{hermans2017defense} by using pre-trained network for feature extraction. The images are resized to 224x224 and have feature extracted with inceptionResnetV2 network \cite{szegedy2017inception} pre-trained with Imagenet dataset. Each image is represented as a 1536-dimensional feature vector. Our trainable extractor network for the feature is simple (1024-128) feed-forward network which reduces data to 1024 dimensions and to 128 dimensions at output layer. For EPID Gamma images, we use same extractor architecture as \cite{nyflot2019deep} which consist of 4 consecutive convolution and max-pooling with filter size 5x5x32 followed by (1024-128) feed-forward layer. Similar network is used as extractor network for Sarcoma scans, which consist of 4 consecutive 3x3 convolution and max-pooling with increasing filter channels of 16,32,64, and 128 respectively. Afterward, the output is sent to (1024-128) feed-forward layer.

We evaluate 2 models of our propose strategy which are local-margin triplet training with and without local positive/negative mining (LM-Mining/ LM). After training, trained extractor is used to extract feature from training and testing set. Then, k-NN classification is done on the testing. We set $k=\ceil{\sqrt{n}}$ where $n$ is number of training instances, which equal to 233,213,17,15 for MNIST, Cifar, EPID, and Sarcoma dataset respectively. Classification results of the proposed models are compared to 3 baselines. First two baselines are typical triplet max-margin training with $m=1,000,000$ with and without hard-negative mining (MM-HardMin/ MM). All models based-on triplet loss are subject to same statistical regularization presented previously. Value of regularization weights are set equally in all method depend on dataset. In all dataset, $c_b=3,w_{ms}=1,w_{md}=1,w_{ss}=0,,w_{lm}=1000$. $w_{sd}$ is set to $1, 0.01, 0 , 0$ for MNIST, Cifar-10, EPID, and Sarcoma respectively The third baseline is training the extractor for end-to-end classification with softmax function and cross-entropy loss. In order of MNIST-Cifar-EPID-Sarcoma datasets, the batch size is set to 128-128-30-20. All training is done without data augmentation except on Sarcoma dataset in which simple augmentation of random flipping vertically and horizontally are applied. In addition to baselines, we also compare our results with a reported state-of-the-art method trained without data augmentation \cite{chang2015batch} for MNIST and Cifar, and \cite{nyflot2019deep} for EPID.

The second experiment is transferring trained feature to other learning model for small-size data to see whether the feature is also useful for other tasks. We use only EPID and Sarcoma datasets in this experiment. For EPID gamma images, the feature is transferred for classification with commonly used simple classifiers for medical tasks. The all trained features are reduced to 32 dimensions with principal component analysis (PCA) since many typical classifiers perform well with low dimensional. After reduction, the features are used to train support vector machine (SVM) with linear kernel, decision tree (DT), multi-layer perceptron (MLP) with 24-dimension hidden layer, ReLu activation, and softmax output, and k-NN with same k setting as that of experiment 1. SVM, DT, and MLP is implemented using scikit-learn package with python. feature of proposed models and baselines are compared using accuracy measure. 

For Sarcoma, the feature is transferred to be used in survival regression. \citeauthor{christ2017survivalnet} explored that feature trained for survival classification can also be used for regression. In our transfer experiment, the trained feature is PCA-reduced to 8 dimensions. Then, the feature is used to trained Cox-proportional hazard model \cite{fox2002cox} which output relative hazard value base on time-until-death. The regression model is implemented using lifeline package. Results are compared using concordance-index (C-Index). Additionally, we also compare results with end-to-end training similar to \cite{zhu2016deep} with same extractor network and same data augmentation for Sarcoma in out experiment.

All networks are trained until converges using Adams optimizer \cite{kingma2014adam} with learning rate set to 0.0001. We observed that Triplet based network converge within 50-60 epoches.
\subsection{Results}

\begin{table}[t]
\caption{Comparison of accuracy(\%) of features from proposed method and baselines in 4 dataset using K-NN classification}
\label{exp1-results}
\begin{center}
\begin{small}
\begin{sc}
\begin{tabular}{lcccc}
\toprule
Models & MNIST & Cifar & Epid & Sarcoma\\
\midrule
(proposed) \\
lm-mining	   & 99.24 & \textbf{92.02} & \textbf{68.24} & 69.44 \\
lm		   & \textbf{99.30} & 91.79 & 65.10 & \textbf{72.22} \\
(baseline)\\
mm		   & 98.63 & 91.45 & \textbf{68.24} & 58.33 \\
mm-hardmin & 81.72 & 91.14 & 41.57 & 55.56 \\
softmax	 & 98.64 & 87.04 & 62.75 & 63.89 \\
\midrule
Best-known	& 99.76 & 93.25 & 64.3 & - \\
\bottomrule
\end{tabular}
\end{sc}
\end{small}
\end{center}
\end{table}

\begin{figure}[t]
\begin{center}
\centerline{\includegraphics[width=0.70\columnwidth]{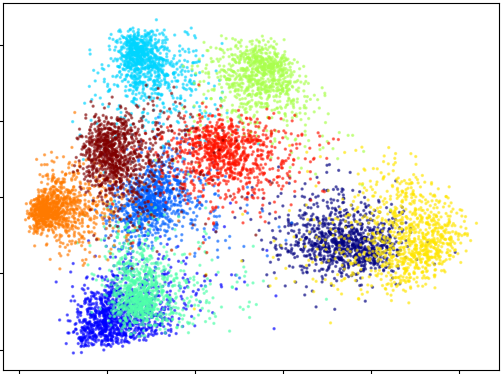}}
\caption{Scatter plot of MNIST feature trained with proposed method (LM-Mining) in after PCA-reduced to 2 dimension.}
\label{MNISTLMMining}
\end{center}
\end{figure}

\begin{figure}[t]
\begin{center}
\centerline{\includegraphics[width=0.70\columnwidth]{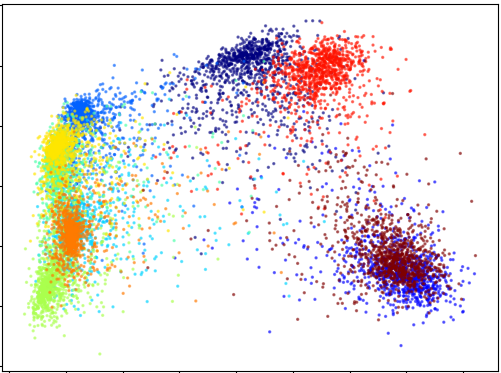}}
\caption{Scatter plot of Cifar10 feature trained with proposed method (LM-Mining) after PCA-reduced to 2 dimension.}
\label{CifarLMMining}
\end{center}
\end{figure}

The results for classification of all dataset is presented in Table~\ref{exp1-results}.  Most of our proposed method outperform baselines using fixed max-margin and softmax especially on small dataset. Even though the propose strategies doesn't performs better than state-of-the-art trained without data augmentation on public data, their results is consistently close within $<1\%$ and $<2\%$ from the-state-of the-art method for MNIST and Cifar-10 respectively. Scatterplots of test features from proposed methods (LM-Mining) for MNIST and Cifar10 in Figure~\ref{MNISTLMMining} and ~\ref{CifarLMMining} also depict class-separation of the feature.


Hard negative mining underperforms in our setting. In later investigation, we find that performance of hard negative/positive mining can improve to be on par with our method of current setting when batch size is 3-4 times of the current size. Nevertheless, it shows limitation of hard negative mining compare to our local-mining strategy with same setting.

End-to-end softmax underperforms, especially on small dataset, compare to our proposed methods shows limitation the loss in comparison to our proposed triplet loss under embedding architecture and dataset. It is also notable that common enhancing techniques commonly used with end-to-end state-of-the-art based on softmax, such as batch-norm and drop-out, are not applied in the experiment. Nevertheless, the results show that our propose method have potential to outperform end-to-end softmax loss in some settings.

The results for transfer learning are in Table~\ref{exp2-results} and~\ref{exp3-results}. For EPID dataset, the feature performs slightly worse for KNN classification compare to full feature before reduction. However, it still consistently outperforms the features trained with softmax baseline and slightly outperform fixed margin baseline in multiple choices of classifiers. The best performance is the fixed margin baseline with SVM. Nevertheless, there is no guarantee whether separation-based metric learning would guarantee improve performance. Thus, we conclude from the results that our transferred feature maintains similar performance and utility compare to that of typical fixed max-margin baseline for transferring to other classification models.

For Sarcoma dataset, one of our features show best performance compare to baseline and achieve C-Index of 0.7624 and largely outperform all baseline. Interestingly, the feature also performs better than end-to-end baseline for survival regression. It is also notable that softmax performs worse than fixed margin despite superior performance in classification. Possibly, unsuitable in tranferring task other than classification. Nevertheless, it is safe to state that the proposed feature can be transfer to related task which is not classification.

\begin{table}[t]
\caption{Comparison of accuracy(\%) of transfer feature with PCA-reduction to 32 dimensions using 4 commonly used classifiers}
\label{exp2-results}
\begin{center}
\begin{small}
\begin{sc}
\begin{tabular}{lcccc}
\toprule
Models & KNN & SVM & DT & MLP\\
\midrule
(proposed) \\
lm-min	   & \textbf{67.45} & 67.45 & \textbf{63.53} & 57.65 \\
lm		   & 65.10 & 65.10 & 62.35 & \textbf{59.61} \\
(baseline)\\
mm		   & 65.49 & \textbf{68.24} & 61.96 & 57.25 \\
softmax	 & 52.16 & 58.04 & 45.88 & 52.55 \\
\bottomrule
\end{tabular}
\end{sc}
\end{small}
\end{center}
\end{table}

\begin{table}[t]
\caption{Comparison of C-Index of Cox survival regression model results of feature from proposed training method and baselines in Sarcoma data}
\label{exp3-results}
\begin{center}
\begin{small}
\begin{sc}
\begin{tabular}{lcccr}
\toprule
Methods  & c-Index \\
\midrule
Lm-Min   & 0.7462\ \\
Lm  & \textbf{0.7624} \\
MM &  0.6786 \\
Softmax  &   0.5256 \\
DeepConvSurv & 0.6197\\
\bottomrule
\end{tabular}
\end{sc}
\end{small}
\end{center}
\end{table}

\section{Conclusion}
We successfully introduce triplet local-margin loss along with its associated local mining strategy. Minimizing the loss results in features with better performance of KNN classification as it improves homogeneity among neighborhoods in feature space. The method reduces the burden of manually tuning and validating subsequent hyper-parameter by integrating NN criteria into triplet loss such that $k$ setting doesn't have to be done by repeated tuning. Our experiments show  the potential to outperform end-to-end softmax and fixed max-margin for kNN classification while retaining similar utility for transferring to other related tasks. Good performance across different models opens the opportunity for wide ranges of further analysis and interpretation of the feature without entirely relying on black-box approach. Thus, we conclude that our proposed method with tuned KNN strategy can serve as a good baseline for deep representation learning for classification especially, in the case of end-to-end method which fail to achieve good performance (e.g. due to small sample/little or no data augmentation.) We envision that the method can help make deep learning models more transparent as KNN is more probabilistically interpretable. Nevertheless, the investigation is a subject for future work. 


\bibliographystyle{aaai}
\bibliography{neighborhoodWatch_revision3_preprint.bbl}
\end{document}